\documentclass[lettersize,journal]{IEEEtran}
\usepackage{amsmath,amsfonts}
\usepackage{algorithmic}
\usepackage{algorithm}
\usepackage{array}
\usepackage[caption=false,font=normalsize,labelfont=sf,textfont=sf]{subfig}
\usepackage{textcomp}
\usepackage{stfloats} 
\usepackage{url}
\usepackage{verbatim}
\usepackage{graphicx}
\usepackage{cite}
\usepackage[hidelinks]{hyperref}
\usepackage{xcolor}

\usepackage{multirow}
\usepackage{makecell}

\DeclareMathOperator*{\argmin}{arg\,min}

\definecolor{ReviseColor}{RGB}{0,0,255}
\definecolor{DefaultColor}{RGB}{0,0,0}

\colorlet{HighlightColor}{DefaultColor}

\begin{document}

\title{EvRepSL: Event-Stream Representation via Self-Supervised Learning for Event-Based Vision}

\author{Qiang Qu, Xiaoming Chen*\thanks{*Corresponding authors}, Yuk Ying Chung, Yiran Shen*,~\IEEEmembership{Senior Member,~IEEE}

        % <-this % stops a space

\IEEEcompsocitemizethanks{
\IEEEcompsocthanksitem Qiang Qu and Yuk Ying Chung are with School of Computer Science, the University of Sydney, Australia. E-mail: vincent.qu@sydney.edu.au; vera.chung@sydney.edu.au
\IEEEcompsocthanksitem Xiaoming Chen is with Beijing Technology and Business University, China.
E-mail:xiaoming.chen@btbu.edu.cn
\IEEEcompsocthanksitem Yiran Shen is with School of Software, Shandong University, China. 
E-mail: yiran.shen@sdu.edu.cn
\IEEEcompsocthanksitem Python Implementation: https://github.com/VincentQQu/EvRepSL
}

% \thanks{This paper was produced by Qiang Qu (vincent.qu@sydney.edu.au), Yuk Ying Chung (vera.chung@sydney.edu.au) in the University of Sydney, Xiaoming Chen (xiaoming.chen@btbu.edu.cn) in the Beijing Technology and Business University, and Yiran Shen (yiran.shen@sdu.edu.cn) in the Shandong University.}% <-this % stops a space
\thanks{Manuscript received October 15, 2023; revised September 22, 2024; accepted 8 November 2024. Date of publication 19 November 2024}}

% The paper headers
\markboth{IEEE TRANSACTIONS ON IMAGE PROCESSING,~Vol.~33, 2024}%
{Shell \MakeLowercase{\textit{et al.}}: A Sample Article Using IEEEtran.cls for IEEE Journals}

% \IEEEpubid{0000--0000/00\$00.00~\copyright~2021 IEEE}
% Remember, if you use this you must call \IEEEpubidadjcol in the second
% column for its text to clear the IEEEpubid mark.

\maketitle

\begin{abstract}
Event-stream representation is the first step for many computer vision tasks using event cameras. It converts the asynchronous event-streams into a formatted structure so that conventional machine learning models can be applied easily. However, most of the state-of-the-art event-stream representations are manually designed and the quality of these representations cannot be guaranteed due to the noisy nature of event-streams.
In this paper, we introduce a data-driven approach aiming at enhancing the quality of event-stream representations. Our approach commences with the introduction of a new event-stream representation based on spatial-temporal statistics, denoted as \textbf{EvRep}.
Subsequently, we theoretically derive the intrinsic relationship between asynchronous event-streams and synchronous video frames.
Building upon this theoretical relationship, we train a representation generator, \textbf{RepGen}, in a self-supervised learning manner accepting \textbf{EvRep} as input.
Finally, the event-streams are converted to high-quality representations, termed as {\bf EvRepSL}, by going through the learned \textbf{RepGen} (without the need of fine-tuning or retraining).
Our methodology is rigorously validated through extensive evaluations on a variety of mainstream event-based classification and optical flow datasets (captured with various types of event cameras). The experimental results highlight not only our approach's superior performance over existing event-stream representations but also its versatility, being agnostic to different event cameras and tasks.
\end{abstract}


\section{Introduction} \label{sec:intro}

\IEEEPARstart{I}{nspired} by human vision system, Silicon Retina \cite{mahowald264mead} has shown a novel and effective form of perceptual sensing, kindling the nascent discipline of neuromorphic engineering. In recent years, this bio-inspired technology has attracted considerable attention from both academia and industry due to the availability of prototype event cameras or Dynamic Vision Sensors (DVS)~\cite{lichtsteiner2008128, berner2013240, posch2010qvga}. Event cameras show a number of unique advantages over active pixel sensors (APS), i.e., the standard frame-based cameras, in a range of computer vision tasks~\cite{delbruck2007fast, gallego2017event, rebecq2019high, jia2023event, sabatier2017asynchronous, cadena2021spade} including visual odometry/SLAM~\cite{kueng2016low, kim2016real, mueggler2017event}, object classification \cite{orchard2015hfirst, lagorce2016hots, sironi2018hats, bi2019graph}, object tracking~\cite{zheng2022spike}, and optical flow estimation~\cite{zhu2018ev, zhu2019unsupervised, gehrig2019end, cannici2020differentiable}. Different from conventional frame-based cameras, which create synchronized frames at predetermined rates, the pixels of event cameras are capable of independently capturing microsecond-level intensity changes and generating a stream of asynchronous ``events''. The unique design of event cameras enables several benefits over frame-based cameras.  First, the temporal resolution of event cameras is up to tens of microseconds, allowing them to record intricate motion phases or high-speed motions without blurring or rolling shutter issues. Second, event cameras enable a substantially higher dynamic range (up to 140dB \cite{lichtsteiner2008128}) than the frame-based cameras (60dB), allowing them to function well under challenging lighting conditions. Third, they consume significantly less resources, including energy, bandwidth, and processing power, since events are sparse and only triggered when intensity changes happen. For instance, DVS128 sensor platform consumes 150 times less energy than a conventional CMOS camera \cite{lichtsteiner2008128}.
These properties make event cameras desirable for computer vision tasks with specific latency, resource, and operating environment constraints.

\begin{figure*}[htbp]
  \centering
%   \fbox{\rule{0pt}{2in} \rule{0.9\linewidth}{0pt}}
   \includegraphics[width=1\linewidth]{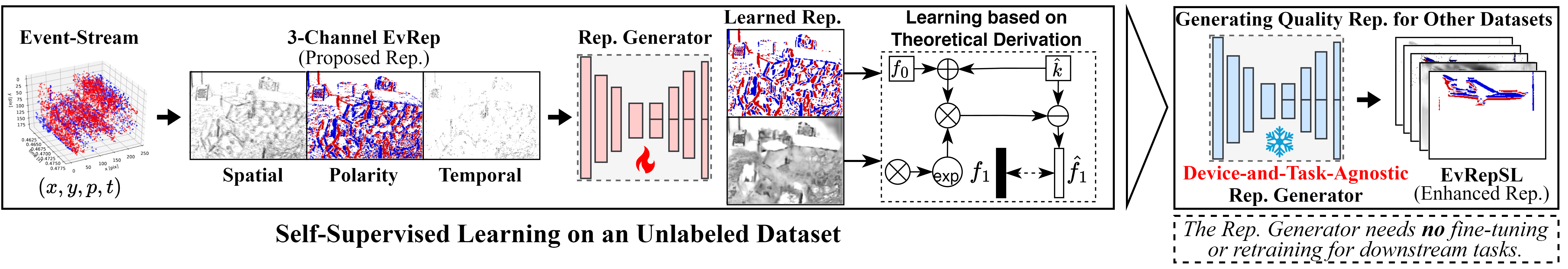}

   \caption{ \textcolor{HighlightColor}{ \textbf{Overview of the Proposed Methodology.} The figure illustrates the self-supervised learning framework for generating event-stream representations. \textbf{Left:} Raw event streams are processed into a 3-channel \textbf{EvRep} representation capturing spatial, polarity, and temporal information. These representations are then used to train the representation generator, \textbf{RepGen}, based on a hybrid, unlabelled dataset. \textbf{Middle:} Through a self-supervised learning process, \textbf{RepGen} learns an enhanced event representation without requiring labeled data. \textbf{Right:} Once trained, \textbf{RepGen} becomes device- and task-agnostic, enabling it to generate high-quality representations (\textbf{EvRepSL}) for other event-only datasets \textit{without requiring fine-tuning or retraining}. This flexibility allows \textbf{EvRepSL} to integrate seamlessly with various downstream event-based tasks, functioning similarly to traditional representations while maintaining its versatility. } }
   \label{fig:trigger}
\end{figure*}

Event cameras produce a stream of events that encode the time, location, and polarity (positive or negative) of brightness changes. Consequently, each individual event conveys little information about the scene. This characteristic of event-streams necessitates the use of integration encoding or representation techniques that aggregate both spatial and temporal information prior to undertaking any event-based applications. Moreover, the unique spatial-temporal structure makes it difficult to be processed using well-researched computer vision algorithms. 
Some researchers address this issue by introducing spiking neural networks (SNNs) to process events as spikings~\cite{lee2016training, orchard2015hfirst, zhao2014feedforward, benosman2013event}. However, the lack of specialized hardware and efficient backpropagation methods limits the use of SNNs. 
Other researchers have developed event-stream representations to convert event-streams into formatted-tensors that are compatible with cutting-edge computer vision methods~\cite{sironi2018hats, maqueda2018event, zhu2018ev, zhu2019unsupervised, gehrig2019end}. Voxel-grid representation~\cite{zhu2019unsupervised}, for instance, represents event-streams in a space-time histogram in which each voxel accounts for a particular pixel and time interval. This representation is compatible with convolutional neural networks (CNNs) and preserves both spatial and temporal information.

However, the majority of existing event-stream representations suffer from the noisy nature of event-streams. Baldwin et al. \cite{baldwin2020event} summarized four types of random noise in event-streams, including background activities (events occurring with no intensity changes), holes (events not occurring with intensity changes), stochastic arrival time of the events, and stochastic number of events for a determined intensity change. They have substantial influence on the precision and bandwidth consumption~\cite{guo2022low, baldwin2020event}. Therefore, before converting event-streams,  some denoising techniques~\cite{wu2020probabilistic, guo2022low, baldwin2020event} could be adopted. However, as there is no groundtruth for real-world eventwise noises (i.e. labels indicating whether an event is a noise or not), efforts on events denoising typically rely on strong assumptions and are ineffective in complex real-world scenario. 

In this work, rather than focusing on eventwise denoising, we introduce an event-stream representation generator, \textbf{RepGen}, that leverages a theory-inspired self-supervised learning framework to directly enhance the quality of event-stream representation, as conceptually depicted in Fig.~\ref{fig:trigger}. Specifically, to preserve the temporal pattern inherent to event-streams, we formulate a novel event-stream representation grounded in spatial-temporal statistics, termed as \textbf{EvRep}. Following this, we theoretically establish the intrinsic relation between asynchronous event-streams and synchronous video frames. Using this established relationship as a foundation, we train our \textbf{RepGen} in a self-supervised manner, taking \textbf{EvRep} as its input. Consequently, event-streams are transformed into superior representations, designated as \textbf{EvRepSL}, via the trained \textbf{RepGen} without subsequent fine-tuning or retraining. Through extensive experiments, we validate that \textbf{EvRepSL} significantly bolsters the performance of event-based tasks.
The contributions of this work are summarized as follow:
\begin{itemize}
    \item We propose {\bf EvRep}, a new event-stream representation based on spatial-temporal statistics to preserve the temporal pattern inherent to event-streams for event-based vision, compatible with the conventional computer vision tasks. 
    \item Based on the theoretical relation between APS frames and event-streams, we design a novel self-supervised learned \textbf{RepGen} to derive high-quality representation, {\bf EvRepSL}, from {\bf EvRep}. The learned \textbf{RepGen} can then generate high-quality representation for event-streams \textit{without the need of frames.}
    \item Through comprehensive evaluations on diverse event-based tasks and datasets, we demonstrate that our proposed \textbf{EvRepSL} can significantly enhance the accuracy of classification and optical flow estimation tasks. Moreover, our experimental results underscore the representation's agnosticism towards varying event cameras and tasks (no fine-tuning is needed), highlighting its versatility. This suggests its potential for broad applicability in various event-based tasks.
\end{itemize}

\section{Related Work}
\label{sec:related_works}

In this section, we review the works related to event-stream representation.
In general, representations of event-streams can be categorized as either event-spikings or formatted-tensors. \textcolor{HighlightColor}{The event-spikings preserve the original property of event-streams (especially asynchrony) and feed to recurrent models like SNN for vision tasks such as object classification~\cite{lee2016training, orchard2015hfirst, perez2013mapping, zhao2014feedforward, barchid2023spiking}, and optical flow estimation~\cite{benosman2013event,benosman2012asynchronous}.} However, the absence of specialized hardware and computationally efficient backpropagation methods still restricts the application of SNNs in complex real-world applications. Recursive methods are another options for event-spikings. For example, phased-LSTM \cite{neil2016phased} adds event timestamps as an additional input to recurrent LSTM, in order to keep asynchronous temporal information. Since events are fed sequentially into models, it is difficult for such models to capture spatial information in long sequences.

\subsection{Event-Stream Representation}
 
The mainstream of event-stream representation is to convert event streams into formatted-tensors \cite{sironi2018hats, maqueda2018event, zhu2018ev, zhu2019unsupervised, gehrig2019end, deng2021learning, ding2022spatio}. Sironi et al. \cite{sironi2018hats} convert events into histograms of averaged time surfaces (HATS) for object classification tasks. Gehrig et al. \cite{gehrig2019end} introduce a learnable event encoder for transforming representations into formatted-tensors. Maqueda et al. \cite{maqueda2018event} employ a simpler representation to address steering prediction, where positive and negative polarity events are stacked into two-channel image-like representation. On the top of two-channel representation, Zhu et al. \cite{zhu2018ev} stack two additional channels containing ratios characterizing temporal properties for optical flow estimation. Then spatial-temporal voxel grid~\cite{zhu2019unsupervised} is proposed for accommodating spatial-temporal information of event-streams. Bi et al. \cite{bi2020graph} achieve state-of-the-art performance in event-based object classification, where graphs are constructed based on events and then go through a graph neural network (GNN). \textcolor{HighlightColor}{Barchid et al. \cite{barchid2022bina} introduce a method that converts asynchronous event streams into sparse, expressive N-bit event frames, achieving state-of-the-art performance and robustness against image corruptions compared to other event representation methods.} Ding et al. \cite{ding2022spatio} propose the Gaussian weighted polarities for optical flow estimation. \textcolor{HighlightColor}{Barchid et al.~\cite{barchid2023exploring} introduce a self-supervised learning framework for event-based vision using lightweight CNNs and novel augmentations, achieving competitive recognition performance and providing insights into feature quality.} Recently, four-channel and voxel-grid have gained popularity \cite{wang2019ev, wang2021event, weng2021event, paredes2021back} because they can be easily implemented, and compatible with the conventional deep learning models. \textcolor{HighlightColor}{There are other work in event-based vision has introduced several representation techniques. Bai et al. \cite{bai2022accurate} presented an accurate and efficient event representation (we denote it as AEER) for object recognition that achieves both accuracy and efficiency by capturing asynchronous events in a structured format. Baldwin et al. \cite{baldwin2022time} proposed Time-Ordered Recent Event (TORE) volumes, which efficiently process event data for pattern recognition tasks by encoding temporal information more effectively. Jiao et al. \cite{jiao2021comparing} explored different event representations in the context of SLAM, comparing their effectiveness in tracking and improving localization accuracy. Liu et al. \cite{liu2023motion} focused on enhancing object detection under motion, developing a lightweight approach that remains robust even at high speeds using event cameras. Additionally, Nunes et al. \cite{nunes2023adaptive} introduced an adaptive global decay process to refine event data, optimizing temporal decay for improved performance in dynamic scenes.}

However, due to the noisy nature of event-streams \cite{wu2020probabilistic, guo2022low, baldwin2020event}, the majority of event-stream representations suffer from the inherited noises. In \cite{baldwin2020event}, Baldwin et al. summarized four forms of random noise present in event-streams. First, an event is triggered when there is no actual change in intensity. These false alarms, often known as background activities, have a significant impact on algorithm precision and use bandwidth. Second, an event is not generated, despite an intensity change (i.e. holes or false-negatives). Third, the arrival time of the event is stochastic. Finally, the actual number of events for a given magnitude fluctuates randomly, despite being proportional to the magnitude of the edge (e.g., a high contrast change generates more events than a low contrast change).

\subsection{\textcolor{HighlightColor}{EfficientNets and UNet}}

\textcolor{HighlightColor}{In this section, we review existing deep learning models to identify a suitable backbone for the proposed self-supervised learning approach, particularly in the context of event representation learning. Due to the nature of event cameras, which are designed for fast and efficient data capture, it is essential to choose an algorithm that mirrors these qualities. Convolutional Neural Networks (CNNs) are generally faster and more efficient than transformer-based architectures, making them a more suitable choice for event-based data processing.}

\textcolor{HighlightColor}{Several CNN architectures have set benchmarks in conventional image classification, including VGGNet \cite{simonyan2014very}, InceptionNet \cite{szegedy2017inception}, ResNet \cite{he2016deep}, MobileNet \cite{howard2017mobilenets, sandler2018mobilenetv2}, SENet \cite{hu2018squeeze}, EfficientNet \cite{tan2019efficientnet}, and the recent EfficientNetV2 \cite{tan2021efficientnetv2}. Among these, EfficientNetV2 \cite{tan2021efficientnetv2} is chosen as the backbone for its superior efficiency in training time and parameter usage, making it particularly well-suited for event representation learning, where speed and resource management are critical.}

\textcolor{HighlightColor}{EfficientNetV2 is developed through training-aware neural architecture search and scaling, resulting in a model that minimizes both training time and computational cost, while maintaining high accuracy. In this work, we leverage the MBConv and Fused MBConv layers from EfficientNetV2 to construct the backbone of the proposed self-supervised learning model, \textbf{RepGen}. The encoder-decoder structure of \textbf{RepGen} follows the general design of UNet \cite{ronneberger2015u}, but with a shared downsampling encoder and two specialized upsampling decoders, optimized for distinct tasks (explained in Section \ref{subsubsec:self_learning_with_APS_DVS}).}

\section{Methodology}
\label{sec:method}

In this section, we will describe the framework of event-stream representation via self-supervised learning in details. It starts with a new representation of event-streams consisting of three statistical channels, termed as {\bf EvRep} (see Section \ref{subsec:evrep}). Based on the imaging principle of event camera~\cite{lichtsteiner2008128}, we theoretically derive a number of key parameters that need to be estimated to improve the quality of the {\bf EvRep} (see Section \ref{subsubsec:theory}). Then, a event-stream representation generator, \textbf{RepGen}, is trained via a self-supervised manner to learn the key parameters obtaining refined representation of the event-streams, i.e., {\bf EvRepSL} (see Section \ref{subsubsec:self_learning_with_APS_DVS}, \ref{subsubsec:repgen}). At last, lightweight deep learning models are designed to accommodate the {\bf EvRepSL} of event-streams for different vision tasks, i.e., classification and optical flow estimation (see Section \ref{subsec:method_classify}, \ref{subsec:method_flow}).

\subsection{The Design of {\bf EvRep}}
\label{subsec:evrep}

\textcolor{HighlightColor}{We propose the initial representation, {\bf EvRep}, directly from the raw event stream to preserve as much event information as possible. It is known that each event contains three essential components: a spatial component $(x, y)$, a polarity component \(p\) (representing the direction of brightness change), and a temporal component \(t\) (the event’s timestamp). To comprehensively retain these three components, we design {\bf EvRep} with three pixel-wise statistical channels: the event spatial channel \((\mathcal{E}_C)\), which captures the spatial information by recording the number of events at each pixel; the event polarity channel \((\mathcal{E}_I)\), which captures the polarity of brightness changes by integrating positive and negative events; and the event temporal channel \((\mathcal{E}_T)\), which preserves the temporal information by accounting for the distribution of events over time.}

\begin{figure}[htbp]
  \centering
%   \fbox{\rule{0pt}{2in} \rule{0.9\linewidth}{0pt}}
   \includegraphics[width=1\linewidth]{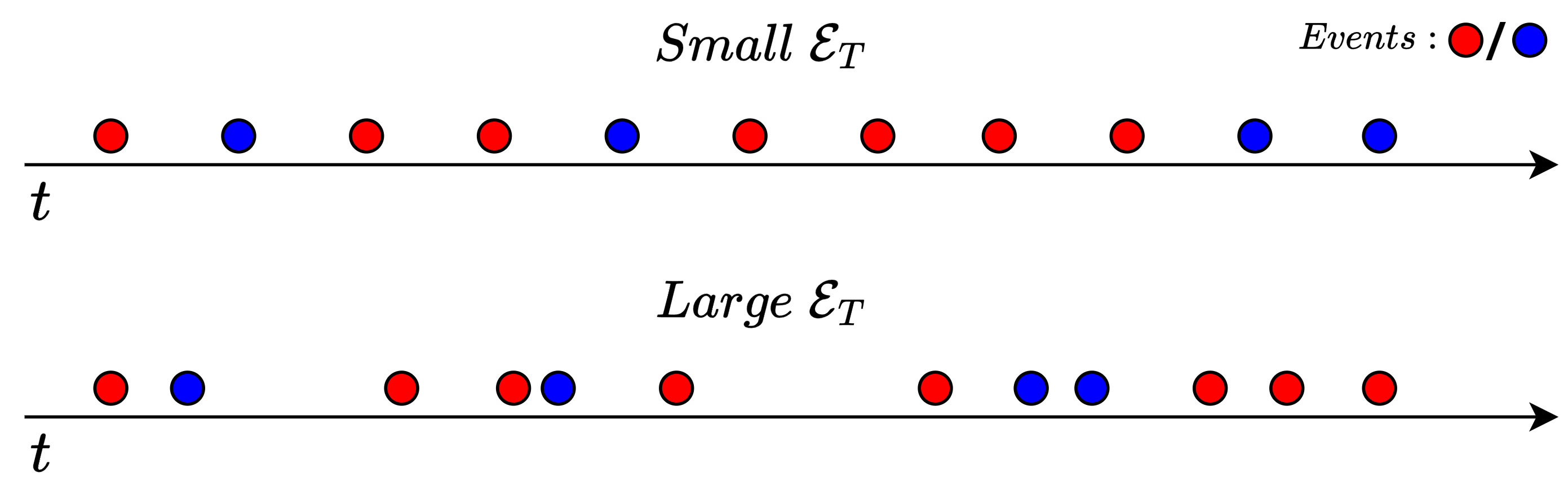}
   \caption{ \textcolor{HighlightColor}{  \textbf{Demonstration of varying event-stream patterns in the temporal domain,} as reflected by different values of the proposed temporal channel $\mathcal{E}_T$, which captures the distinctive timing dynamics of the events. } }
   \label{fig:temporal_std}
\end{figure}

Specifically, in the time interval $[t_0, t_1]$, $n_{x,y}$ events $\{e_i^{x,y}\}_0^{n_{x,y}-1}$, are generated at the pixel $(x, y)$ from a $w \times h$ image plane. Then the event spatial channel $\mathcal{E}_C$ at pixel $(x,y)$ is:

\begin{equation}
\label{eq:def_tc}
% \begin{aligned}
\mathcal{E}_C(x,y) = n_{x,y}.
% \end{aligned}
\end{equation}
The event polarity channel $\mathcal{E}_I$ is calculated by integrating the events with polarities at any pixel $(x,y)$:

\textcolor{HighlightColor}{\begin{equation}
\label{eq:def_ie}
% \begin{aligned}
\mathcal{E}_I(x,y) = \sum_{i=0}^{n_{x,y}-1}p_i^{x,y}, \quad where \quad p \in\{-1, 1\}.
% \end{aligned}
\end{equation}}
Intuitively, $\mathcal{E}_I$ measures the \textit{net numbers of events} at each pixel by cancelling out the same number of positive and negative events. 
\textcolor{HighlightColor}{At last, the event temporal channel $\mathcal{E}_T$ is holding the temporal statistics of the event-streams. The timestamp of the $i_{th}$ event at pixel $(x, y)$ is $\tau^{x,y}_i$ and the time difference between each pair of consecutive events is:}
\textcolor{HighlightColor}{\begin{equation}
    \delta_{i}^{x,y}=\tau^{x,y}_{i+1}- \tau^{x,y}_{i}.
\end{equation}
Then $\mathcal{E}_T$ at pixel $(x, y)$ can be expressed as:
\begin{equation}
\label{eq:def_et}
\begin{aligned}
\mathcal{E}_T(x,y) = \sqrt{ \frac{ \sum_{i=0}^{n_{x,y}-2} (\delta_i^{x,y} - \delta^{x,y}_{mean})^2 }{ n_{x,y}-1 } },
\end{aligned}
\end{equation}
where $\delta^{x,y}_{mean} = \sum_{n=0}^{n_{x,y}-2}\delta_i^{x,y}/n_{x,y}$. We use the sample standard deviation (with $n_{x,y} - 1$ in the denominator) rather than the population standard deviation to account for the fact that the observed events at each pixel $(x, y)$ are treated as a sample from a potentially continuous and larger event stream. 
Therefore, the channel $\mathcal{E}_T$ actually accounts for the standard deviation of the time difference between consecutive events.}
It measures the occurring pattern of the event-stream in temporal domain. For example, if $\mathcal{E}_T^{x,y}=0$, events occurs uniformly over time, and larger $\mathcal{E}_T^{x,y}$ indicates events are generated more dynamically (as demonstrated in Fig.~\ref{fig:temporal_std}). Therefore, $\mathcal{E}_T$ can effectively model the temporal information of event-streams.

Finally, {\bf EvRep} is defined as:
\begin{equation}
  EvRep = \{\mathcal{E}_I, \mathcal{E}_C, \mathcal{E}_T\}.
  \label{eq:evrep}
\end{equation}

\textcolor{HighlightColor}{The design of {\bf EvRep} effectively captures the key components of event streams—spatial, polarity, and temporal information—through its three statistical channels: event count, polarity, and temporal distribution. This comprehensive approach preserves critical data, ensuring a robust foundation for downstream tasks that require detailed spatiotemporal and polarity information. As a result, {\bf EvRep} provides a versatile and efficient representation for event-based data processing.}

\subsection{Event-stream Representation via Self-supervised Learning}

\begin{figure*}[!htb]
  \centering
   \includegraphics[width=1\linewidth]{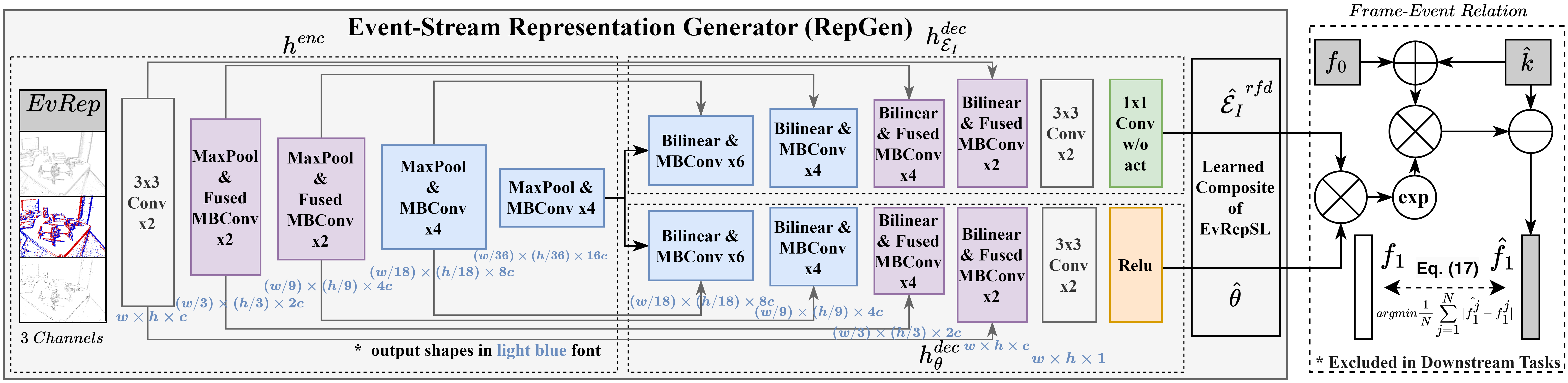}
   \caption{ \textcolor{HighlightColor}{ {\bf Network architecture of the proposed RepGen for self-supervised representation learning.} 
 It consists of a shared downsampling encoder $h^{enc}$ and two separate upsampling decoders ($h^{dec}_{\mathcal{E}_I}$ and $h^{dec}_{\theta}$) exquisitely designed for estimating $\mathcal{E_I}$ and $\theta$ respectively, followed by a computing module to predict the next frame $f_1$. It is important to emphasize that the block Frame-Event Relation is used solely to guide the self-supervised learning process and is not included in downstream applications. In other words, once \textbf{RepGen} is trained during self-supervised learning, it can be directly applied to any event-only datasets, without requiring the presence of frame data.}
 }
   \label{fig:model}
\end{figure*}

Event-streams are normally considered as noisy due to the unstable hardware and lighting condition \cite{wu2020probabilistic, guo2022low, baldwin2020event}. Therefore, {\bf EvRep} from the statistical properties of the event-streams are inherently low-quality. Conventional approaches seek to suppress the noises in the event-streams directly~\cite{baldwin2020event,feng2020event,duan2021eventzoom}, however, the performance of noise reduction is limited by the issues on obtaining the accurate eventwise groundtruth under practical settings.
Instead of denoising the event-streams, we propose a self-supervised learning strategy on  {\bf EvRep} directly to improve its representation quality. Considering the fact that a number of event cameras can generate both event-streams and APS  frames simultaneously, we exploit the theoretical correlation between the event-streams and APS frames to refine the event-stream representation in a self-supervised learning approach to derive {\bf EvRepSL}. 
\subsubsection{Theoretical Derivation}
\label{subsubsec:theory}
Before introducing the detailed design of the self-supervised learning framework, we first derive the theoretical relation between the event-streams and APS frames.  
According to the principal of event camera~\cite{lichtsteiner2008128}, an event is generated when an log-intensity change is detected (over threshold), i.e.,

\begin{equation}
  d_i^{x,y} = log I_{i+1}^{x,y} - log I_i^{x,y}.
  \label{eq:d_i}
\end{equation}
where $I_{i}^{x,y}$ and $I_{i+1}^{x,y}$ are the intensity values of two consecutive events generated at pixel $(x,y)$. The accumulated log-intensity change within $[t_0, t_1]$ (time interval between two consecutive video frames $f_0$ and $f_1$) can be approximated as,

\begin{equation}
\label{eq:sum_d_i}
\begin{aligned}
\sum_{i=0}^{n_{x,y}-1} d_i^{x,y} = log I_n^{x,y} - log I_{n-1}^{x,y} + log I_{n-1}^{x,y}\\ \quad ... - log I_1^{x,y} + log I_1^{x,y} - log I_0^{x,y}.
\end{aligned}
\end{equation}

By cancelling out the same items, the equation above can be rewritten as,
\begin{equation}
  \sum_{n=0}^{n_{x,y}-1} d_i^{x,y} = log I_n^{x,y} - log I_0^{x,y}.
  \label{eq:sum_d_i_clean}
\end{equation}

Then $d_i^{x,y}$ can be decomposed as,

\begin{equation}
\label{eq:d_i_sub}
\begin{aligned}
  d_i^{x,y} = sign_i^{x,y} * {\theta}_i^{x,y},
\end{aligned}
\end{equation}
where $sign_i^{x,y} \in \{-1, 1\}$. ${\theta_i^{x,y}}$ is the absolute value of the threshold for producing an event at pixel $(x,y)$.
When $[t_0, t_1]$ is the interval between two consecutive frames,  we assume $\theta^{x,y}$ is a constant threshold for the pixel $p_{x,y}$ within this short period (normally tens of milliseconds).

For event-streams, $sign_i^{x,y}$ corresponds to the polarity of the event $e_i^{x,y}$, so the integral channel $\mathcal{E}_I$ of {\bf EvRep}  within $[t_0, t_1]$ can be reformulated as:

\begin{equation}
  \mathcal{E}_I(x,y) = \sum_{i=0}^{n_{x,y}-1}sign_i^{x,y}.
  \label{eq:ie}
\end{equation}

With equation Eq.~\eqref{eq:ie}, the accumulative log-intensity change within $[t_0, t_1]$ can be expressed as,

\begin{equation}
  \sum_{i=0}^{n_{x,y}-1} d_i^{x,y} = {\theta}^{x,y} \sum_{i=0}^{n_{x,y}-1}sign_i^{x,y} = {\theta}^{x,y} * \mathcal{E}_I(x,y).
  \label{eq:sum_d_i_c}
\end{equation}

By incorporating Eq.~\eqref{eq:sum_d_i_clean} into Eq.~\eqref{eq:sum_d_i_c}, we have

\begin{equation}
  {\theta}^{x,y} * \mathcal{E}_I(x,y) = log \frac{I_n^{x,y}}{I_0^{x,y}}.
  \label{eq:c_ie}
\end{equation}

Then we assume a linear relationship between intensity $I$ and the corresponding normalized frame $f$, i.e.,

\begin{equation}
\label{eq:I_i}
\begin{aligned}
  I = af + b,
\end{aligned}
\end{equation}
where $a > 0$ and the pixel values of $f(x,y) \in [0, 1]$.
Then,

\begin{equation}
\label{eq:I_n_I_0}
\begin{aligned}
  \frac{I_n^{x,y}}{I_0^{x,y}} = \frac{af_{t_1}^{x,y} + b}{af_{t_0}^{x,y} + b} = \frac{f_{t_1}^{x,y} + k}{f_{t_0}^{x,y} + k},
\end{aligned}
\end{equation}
where $k = b/a$; $f_{t_0}$ and $f_{t_1}$ are normalized frames recorded at timestamps $t_0$ and $t_1$.
According to Eq.~\eqref{eq:c_ie}, we have 

\begin{equation}
  \frac{f_{t_1}^{x,y} + k}{f_{t_0}^{x,y} + k} = exp({\theta}^{x,y} * {\mathcal{E}_I}^{x,y}).
  \label{eq:f_n_f_0}
\end{equation}
Finally the APS frame $f_{t_1}$ can be represented as:
\begin{equation}
  f_{t_1}^{x,y} = exp({\theta^{x,y} * {\mathcal{E}_I}^{x,y})} * (f_{t_0}^{x,y} + k) - k.
  \label{eq:f_n}
\end{equation}

\textcolor{HighlightColor}{According to the equation above, we can establish the relationship between consecutive frames \(f_0\) and \(f_1\), and the channel \(\mathcal{E}_I\) based on the parameters \(\{\theta, k\}\). At a high level, \(\{\theta, k\}\) can be interpreted as an estimation of pixel-wise contrast thresholds, where \(k\) serves as a learnable parameter that maps the intensity values derived from events to traditional vision intensity values. While \(k\) remains constant once trained, \(\theta\) is designed to adapt dynamically to events across different time frames.}
As a number of popular used event cameras (e.g., DAVIS240, DAVIS346, and Celex) can produce both video frames and event-streams simultaneously, it is convenient to obtain such types of datasets publicly, e.g., ~\cite{scheerlinck2019ced, mueggler2017event, stoffregen2020reducing, binas2017ddd17, zhu2018multivehicle}. Therefore, in this paper, we propose a data-driven approach to estimate the parameters above, then the video frames along with the estimated parameters work together to improve the quality of the event-stream representation.

 \subsubsection{Self-Supervised Representation Learning}
\label{subsubsec:self_learning_with_APS_DVS}

According to the relation between frames and events derived above, we propose {\bf RepGen}, trained with a self-supervised learning to estimate the key parameters and improve the quality of the event-stream representation. {\bf RepGen} can be trained with any event-based datasets containing both video frames and event-streams, e.g., IJRR~\cite{mueggler2017event}, HQF~\cite{stoffregen2020reducing}, DDD17~\cite{binas2017ddd17} collected by different hybrid frame-event platforms. Then the trained {\bf RepGen} is a device-and-task-agnostic representation enhancer compatible with any event-based vision task without further fine-tuning.

As shown in Fig.~\ref{fig:model}, {\bf RepGen} starts with taking the {\bf EvRep} = \{$\mathcal{E}_I$, $\mathcal{E}_C$, $\mathcal{E}_T$\} as input to the encoder $h^{enc}$. {\bf EvRep} is obtained from accumulating the events between the two consecutive frames $f_0$ and $f_1$. The encoder consists of multiple convolutional and (Fused) MBConv~\cite{sandler2018mobilenetv2, tan2021efficientnetv2} layers followed by MaxPooling for feature extraction.  The MBConv utilizes the inverted bottleneck structure \cite{sandler2018mobilenetv2} and depthwise convolutional layers \cite{howard2017mobilenets} to improve memory efficiency. In addition, a squeeze-and-excitation unit \cite{hu2018squeeze} is inserted in the middle of MBConv in order to adaptively recalibrate channel-wise feature responses. The fused version of MBConv substitutes the depthwise convolutional layers with regular convolutional layers, which has been shown to be more effective in large spatial dimensions \cite{tan2021efficientnetv2}. Then two non-sharing decoders $h^{dec}_{\mathcal{E}_I}$ and $h^{dec}_{\theta}$ with the same network architecture (except for the last layer as shown in Fig.~\ref{fig:model}) are adopted to decode the low-dimensional features with upsampling layers to obtain  the estimates $\hat{\mathcal{E}_I}^{rfd}$ and $\hat\theta$. The decoder consists of the blocks of combination of spatial bilinear interpolation and (Fused) MBConv layers. A $1\times 1$ convolutional layer without activation function is applied to finalize $h^{dec}_{\mathcal{E}_I}$ and Relu activation function is adopted to ensure the non-negative property of $\theta$. At last, according to Eq.~\eqref{eq:f_n}, the estimate of the next frame, $\hat{f_1}$, can be outputted with estimated $\{\hat{\mathcal{E}_I}^{rfd}, \hat\theta\}$ and known $f_0$.

As $f_1$ is known in the self-supervised learning phrase, {\bf RepGen} can be trained according to the estimation error of $f_1$. The learning objective is then defined as:

\begin{equation}
\label{eq:obj}
\begin{aligned}
\{ h^{enc}, \; h^{dec}_{\mathcal{E}_I}, \; h^{dec}_{\theta} \} &= \argmin {\frac{1}{N}\sum_{j=1}^{N} |\hat{f_1^j}-f_1^j|},
\end{aligned}
\end{equation}

where $N$ is the total number of video frames and the loss function is the Mean Absolute Error (MAE) between the estimate and the actual frame $f_1$.

The trained {\bf RepGen} can generate two more channels of representation with {\bf EvRep} as the input. The first one is a learned integral of events, i.e., $\hat{\mathcal{E}}_I^{rfd}$ and the second one is the learned threshold for each pixel, i.e., $\hat{\theta}$. Together with {\bf EvRep}, the self-learned new representation {\bf EvRepSL} is defined as:

\begin{equation}
  EvRepSL = \{\mathcal{E}_I, \mathcal{E}_C, \mathcal{E}_T, \hat{\mathcal{E}}_I^{rfd}, \hat{\theta}\}.
  \label{eq:evrepsl}
\end{equation}

\subsubsection{Detailed Design of RepGen}
\label{subsubsec:repgen}
Inspired by EfficientNetv2~\cite{tan2021efficientnetv2} and UNet~\cite{ronneberger2015u}, we design an efficient W-shape network for \textbf{RepGen}. The detailed design of each layer is displayed in purple font in Fig.~\ref{fig:model}, where $w \times h$ is the event-stream resolution and $c$ is a hyper-parameter that determines the model complexity. \textcolor{HighlightColor}{The detailed parameter settings of all model layers are also outlined in Table \ref{tab:model_shapes}.}
\begin{table}[htb]
% \small
\scriptsize
\renewcommand{\arraystretch}{1.2}
\caption{\textcolor{HighlightColor}{ {\bf A summary of parameter settings of all layers of RepGen.} The input size is $w \times h \times c_0$, and the number of channels of the first 3x3 Conv filter is $c$. } }
\label{tab:model_shapes}
\centering
\setlength{\tabcolsep}{3pt}
\textcolor{HighlightColor}{
\begin{tabular}{|p{0.17\textwidth}|p{0.1\textwidth}|p{0.16\textwidth}|} \hline
\textbf{Type / Repeat}& \textbf{Filter Shape}& \textbf{Input Size} \\ \hline \hline
\multicolumn{3}{|c|}{Shared Encoder $\;h^{enc}$}\\ \hline
3x3 Conv x2& $ 3 \times 3 \times c $ & $w\times h \times c_0$ \\ \hline
MaxPool, Fused MBConv x2& $ 3 \times 3 \times 2c $ & $w\times h \times c$ \\ \hline
MaxPool, Fused MBConv x2& $ 3 \times 3 \times 4c $ & $(w/3)\times (h/3) \times 2c$ \\ \hline
MaxPool, MBConv x4& $ 3 \times 3 \times 8c $ & $(w/9)\times (h/9) \times 4c$ \\ \hline
MaxPool, MBConv x4& $ 3 \times 3 \times 16c $ & $(w/18)\times (h/18) \times 8c$ \\ \hline
Output Deep Features & $-$ & $(w/36)\times (h/36) \times 16c$ \\ \hline \hline
\multicolumn{3}{|c|}{Decoder $\;h^{dec}_{\mathcal{E}_I}$}\\ \hline
Bilinear, MBConv x6 & $ 3 \times 3 \times 8c $ & $(w/36)\times (h/36) \times 16c$ \\ \hline
Bilinear, MBConv x4 & $ 3 \times 3 \times 4c $ & $(w/18)\times (h/18) \times 8c$ \\ \hline
Bilinear, Fused MBConv x4 & $ 3 \times 3 \times 2c $ & $(w/9)\times (h/9) \times 4c$ \\ \hline
Bilinear, Fused MBConv x2 & $ 3 \times 3 \times c $ & $(w/3)\times (h/3) \times 2c$ \\ \hline
3x3 Conv x2 & $ 3 \times 3 \times c $ & $w\times h \times c$ \\ \hline
1x1 Conv & $ 1 \times 1 \times c $ & $w\times h \times c$ \\ \hline
Output $\;\hat{\mathcal{E}_I}^{rfd}$ & $-$ & $w\times h \times c_0$ \\ \hline \hline
\multicolumn{3}{|c|}{Decoder $\;h^{dec}_{\theta}$}\\ \hline
Bilinear, MBConv x6 & $ 3 \times 3 \times 8c $ & $(w/36)\times (h/36) \times 16c$ \\ \hline
Bilinear, MBConv x4 & $ 3 \times 3 \times 4c $ & $(w/18)\times (h/18) \times 8c$ \\ \hline
Bilinear, Fused MBConv x4 & $ 3 \times 3 \times 2c $ & $(w/9)\times (h/9) \times 4c$ \\ \hline
Bilinear, Fused MBConv x2 & $ 3 \times 3 \times c $ & $(w/3)\times (h/3) \times 2c$ \\ \hline
3x3 Conv x2 & $ 3 \times 3 \times c $ & $w\times h \times c$ \\ \hline
ReLU & $-$ & $w\times h \times c$ \\ \hline
Output $\;\hat{\theta}$ & $-$ & $w\times h \times c_0$ \\ \hline \hline
\multicolumn{3}{|c|}{$f_1\;$ Estimation}\\ \hline
Calculate Intermediate Value & $-$ & $w\times h \times c_0$ \\ \hline
$-\hat{k}$ & $1$ & $w\times h \times c_0$ \\ \hline
Output $\;f_1$ & $-$ & $w\times h \times c_0$ \\ \hline
\end{tabular}
}
\end{table}

Instead of conventional batch normalization after each convolutional layer, we adopt layer normalization operation~\cite{ba2016layer} to allow small batch size in training. Swish~\cite{ramachandran2017searching} is selected as the default activation function to avoid pre-activations:
\begin{equation}
% \small
  \label{eq:swish}
  Swish(x) = \frac{x}{1 + e^{-x}},
\end{equation}
\noindent where $x$ represents input features.

\subsection{Classification with EvRepSL}
\label{subsec:method_classify}
Once {\bf RepGen} has been trained in self-supervised learning, it is capable of converting any event-stream collected by different event cameras to {\bf EvRepSL}. For downstream tasks, the converted {\bf EvRepSL} as a formatted tensor can be readily feed to deep neural networks. It is worth noting that, different from other event-based representation learning approaches, e.g., EST~\cite{gehrig2019end}, the weights of {\bf RepGen} are need not to be fine-tuned to adapt to the downstream tasks. Therefore, the use of {\bf RepGen} saves significant training cost. Object classification is selected as the primary event-based task to validate our proposed method, given its foundational significance in event-based vision~\cite{neil2016phased,sironi2018hats,gallego2020event}. In the classification task, we adopt the same network architecture as the encoder of {\bf RepGen} as the backbone and a classification block as the tail, termed as {\bf Eff-Classifier}. The classification block consists of three layers of $1\times 1$ convolutional layers and a global average pooling layer. Softmax is applied to obtain the final classification results.
\textcolor{HighlightColor}{
The loss function of multi-class cross entropy \cite{cox1958regression} is applied to adapt to the classification task:
\begin{equation}
\ell_{cls} = - \sum_{i=1}^{N} \sum_{c=1}^{C} y_{i,c} \log(\hat{y}_{i,c}),
\end{equation}
}
\textcolor{HighlightColor}{
\noindent where $N$ represents the number of samples, and $C$ is the number of classes. The term $y_{i,c}$ is a binary indicator, where $y_{i,c} = 1$ if class $c$ is the correct classification for sample $i$, and $y_{i,c} = 0$ otherwise. The term $\hat{y}_{i,c}$ denotes the predicted probability that sample $i$ belongs to class $c$. The loss function computes the sum of the negative log probabilities of the correct class for each sample, providing a measure of how well the predicted probabilities match the true labels.
}

\textcolor{HighlightColor}{
Note that the weights of \textbf{RepGen} are frozen after self-supervised learning, allowing {\bf EvRepSL} to be used for downstream tasks without requiring fine-tuning. This versatility enables {\bf EvRepSL} to function similarly to traditional representations, seamlessly integrating with various tasks. Our experiments show that, even with a minimal downstream model {\bf Eff-Classifier}, {\bf EvRepSL} outperforms significantly larger models that rely on existing representations, proving its capacity to boost performance in diverse architectures.
Since {\bf EvRepSL} is a structured tensor, it is compatible with any traditional classifiers. In our experiments, we utilize a very lightweight model, {\bf Eff-Classifier}, to demonstrate its effectiveness. Additionally, we include results from a larger model, ResNet50~\cite{he2016deep}, in the ablation study to validate whether the proposed representation can enhance performance across different model sizes.
}

\subsection{Optical Flow Estimation with EvRepSL}
\label{subsec:method_flow}
Optical flow estimation is selected as the second event-based task to validate our proposed method. Contrasting with object classification, which seeks a high-level understanding of event-streams, optical flow estimation offers a low-level, pixel-wise analysis, representing a distinct facet of event-based tasks. Once \textbf{EvRepSL} is obtained from \textbf{RepGen}, it can also be  utilized to generate high quality input for optical flow estimation. 
We design \textbf{Eff-FlowNet} by adopting similar network design as \textbf{RepGen} as backbone, but with only one decoder which outputs the two-dimensional optical flow for all pixels.
\textcolor{HighlightColor}{
The robust Charbonnier loss \cite{sun2014quantitative} is used as the loss function with the same setting in~\cite{gehrig2019end}:
\begin{equation}
\ell_{flow} = \sum_{i} \left [ \left (\hat{\mathbf{u}}_i - \mathbf{u}_i \right )^2 + \epsilon^2 \right ]^\alpha,
\end{equation}
}

\textcolor{HighlightColor}{
\noindent where $\hat{\mathbf{u}}_i$ represents the predicted optical flow at the $i$-th pixel, and $\mathbf{u}_i$ denotes the ground truth optical flow at the same pixel. The term $\epsilon$ is a small constant added for numerical stability. In our experiments, $\epsilon = 10^{-3}$ and $\alpha = 0.5$, following the setting in~\cite{gehrig2019end}. The Charbonnier loss is calculated by taking the squared difference between the predicted and ground truth optical flow values, adding $\epsilon^2$, and then computing the square root.
}

\textcolor{HighlightColor}{
It is important to emphasize that after the self-supervised learning phase, the weights of RepGen are frozen, allowing {\bf EvRepSL} to be directly utilized for downstream tasks without the need for fine-tuning. This inherent versatility enables {\bf EvRepSL} to function like traditional representations, seamlessly integrating into a variety of tasks. Our experiments further illustrate that even when used with a lightweight downstream model, \textbf{Eff-FlowNet}, {\bf EvRepSL} outperforms significantly larger models that rely on existing representations, underscoring its ability to substantially boost performance across different architectures.
}

\section{Experiments}

\begin{table*}[htbp]
% \begin{tabularx*}{\textwidth}{X|l}
% \captionsetup{font=small}
\caption{{\bf Quantitative comparisons (Top-1 accuracy) of the evaluated event-based classification models} on N-MNIST, N-Caltech101, N-CARS, CIFAR10-DVS, and ASL-DVS. The best results are in bold while the second best results are underlined.}
\label{tab:classification_results}
% \scriptsize
\small
\centering
\renewcommand{\arraystretch}{1.2}
\setlength{\tabcolsep}{2pt}
\begin{tabular}{c|c|ccccc}
\hline
\hline
% \hline
\textbf{Representation Type} & \textbf{Model} & \textbf{N-MNIST~\cite{orchard2015converting}} &\textbf{N-Caltech101~\cite{orchard2015converting}} & \textbf{N-CARS~\cite{sironi2018hats}} & \textbf{CIFAR10-DVS~\cite{li2017cifar10}} & \textbf{ASL-DVS~\cite{bi2019graph}}   \\
% \hline
\hline
Raw Events &Gabor-SNN \cite{lee2016training, sironi2018hats}  & $0.837$ & $0.196$ & $0.789$ & $0.245$ & $-$  \\
\hline
\multirow{2}{*}{Time Surfaces \cite{lagorce2016hots, sironi2018hats}}&HOTS \cite{lagorce2016hots}  & $0.808$ & $0.210$ & $0.624$ & $0.271$ & $-$  \\
&HATS \cite{sironi2018hats}  & \boldmath $0.991$ & $0.642$ & $0.902$ & $0.524$ & $-$  \\
\hline
\multirow{6}{*}{Graph\cite{bi2019graph}}&GCN \cite{kipf2016semi}  & $0.781$ & $0.530$ & $0.827$ & $0.418$ & $0.811$  \\
&ChebConv \cite{defferrard2016convolutional} & $0.949$ & $0.524$ & $0.855$ & $0.452$ & $0.317$  \\
&MoNet \cite{monti2017geometric} & $0.965$ & $0.571$ & $0.854$ & $0.476$ & $0.867$  \\
&GIN \cite{xu2018powerful} & $0.754$ & $0.476$ & $0.846$ & $0.423$ & $0.514$  \\
&G-CNN \cite{bi2020graph}&  $0.985$ & $0.630$ & $0.902$ & $0.515$ & $0.875$  \\
&RG-CNN \cite{bi2020graph}&  $\underline{0.990}$ & $0.657$ & $0.914$ & $0.540$ & $0.901$ \\
\hline
\multirow{3}{*}{Two-Channel \cite{maqueda2018event}}&VGG-19 \cite{simonyan2014very} &  $0.972$ & $0.549$ & $0.728$ & $0.334$ & $0.806$\\
&Inception-V4 \cite{szegedy2017inception} & $0.973$ & $0.578$ & $0.864$ & $0.379$ & $0.832$  \\
& ResNet-50 \cite{he2016deep}& $0.984$ & $0.637$ & $0.903$ & $0.558$ & $0.886$  \\
\hline
Two-Channel \cite{maqueda2018event}& \multirow{5}{*}{Eff-Classifier} & $0.979$ & $0.725$ & $0.918$ & $0.559$ & $0.940$ \\
Four-Channel \cite{zhu2018ev}& & $0.980$ & $0.767$ & $0.920$ & $0.562$ & $0.944$ \\
% \hline
Voxel-Grid \cite{zhu2019unsupervised}& & $0.981$ & $0.765$ & $0.920$ & $0.572$ & $0.964$  \\
EvRep &  & $\underline{0.990}$ & $\underline{0.786}$ & $\underline{0.923}$ & $\underline{0.590}$ & $\underline{0.977}$ \\
EvRepSL & & \boldmath$0.991$ & \boldmath$0.864$ & \boldmath$0.961$ & \boldmath$0.771$ & \boldmath$0.986$ \\

\hline
\hline
\end{tabular}
% \end{tabularx}
\end{table*}

In this section, we will first describe how the {\bf RepGen} is trained with hybrid frame-event dataset (IJRR~\cite{mueggler2017event}) in a self-supervised learning approach. Then we evaluate and compare the performance of our proposed {\bf EvRepSL} with other  event-stream representations and methods on different datasets for classification or optical flow estimation.

\subsection{Self-Supervised Representation Learning}

We train {\bf RepGen} using IJRR \cite{mueggler2017event} dataset in a self-supervised learning approach. IJRR dataset is collected by DAVIS240C event camera which outputs both video frames and event-streams simultaneously. The frame rate of video is approximately 24fps and the spatial resolution of the pixel array, for both video frame and event-stream, is $240\times180$. IJRR contains 27 subsets and each subset records different scenes with different speeds of movement and various poses. It is worth noting that IJRR dataset is only used to train the {\bf RepGen} network for representation refinement and is irrelevant to the datasets used for evaluating its performance on different vision tasks.

When training {\bf RepGen} with IJRR dataset, data augmentation techniques are applied to alleviate the problem of overfitting. First, frames and event-streams (in {\bf EvRep} representation) are randomly rotated by 90, 180 or 270 degrees. Then different frame rates (24fps, 12fps and 8fps) are adopted  by skipping some of the frames evenly.  Finally, model is also trained with frames and events in reversed order to improve the consistency and interpretability of the proposed approach since Eq. \ref{eq:f_n_f_0} can be rewritten as:

\begin{equation}
  f_{t_0}^{x,y} = exp[{\theta^{x,y} * (-{\mathcal{E}_I}^{x,y})]} * (f_{t_1}^{x,y} + k) - k.
  \label{eq:f_0}
\end{equation}

All experiments are conducted on a system equipped with an AMD 5950X processor, RTX 3090 graphics card, and 32GB RAM, running on Ubuntu 20.04 with the PyTorch framework \cite{NEURIPS2019_9015}. 80\% of the frame pairs and the event-streams between them are randomly selected for training and the rest is used for validation. We train the model using the Adam optimizer~\cite{kingma2014adam} for 300 epoch with batches of size 16. The learning rate starts with 10e-3 and decreases by a factor of 10 for every 100 epoch.

\subsection{Evaluation for Event-based Classification Task }

\subsubsection{Evaluation Setup}
\label{subsubsec:cls_training_config}
In this section, we evaluate and compare our proposed {\bf EvRep} and {\bf EvRepSL} with SOTA event-stream representations on five datasets for classification, including N-MNIST \cite{orchard2015converting}, N-Caltech101 \cite{orchard2015converting}, N-CARS \cite{sironi2018hats}, CIFAR10-DVS \cite{li2017cifar10}, and ASL-DVS \cite{bi2020graph}. During the evaluation, we follow the same protocol as in~\cite{bi2020graph} for split of training and testing portions, i.e., 80\% of the data samples are randomly selected as training data and the rest are used as testing data. For the N-MNIST and N-CARS, we use the predefined training-and-testing splits.  N-MNIST and ASL-DVS are trained by 10 epochs with learning rate 1e-3, while for CIFAR10-DVS, and N-CARS, the number of training epochs is set as 100 with learning rate starting from 1e-4 and decreasing by a factor of 0.5 for every 25 epochs. The size of the training batch is set to be 16 for all datasets. As the number of classes in N-Caltech101 is large, 300 training epochs are adopted to alleviate the problem of underfitting. 
In the evaluation, we first compare our proposed event-based classification framework with a number of different SOTA event-based classification methods as shown in Table~\ref{tab:classification_results}.  To demonstrate our proposed representations really take effect, we train {\bf Eff-Classifier} using the two-channel representation~\cite{maqueda2018event}, four-channel representation~\cite{zhu2018ev}, and voxel-grid representation~\cite{zhu2019unsupervised} as input layers. 
At last, all the experiments are repeated for 10 times and the average of the results will be reported.

\subsubsection{Evaluation Results of Classification Task}

\noindent{\bf Comparison on Existing Classification Methods.} We compare the Top-1 accuracy of our classification framework to other event-based classifiers and the results are shown in Table~\ref{tab:classification_results}. The names of the classifiers are shown in the {\bf Model} column and their corresponding representations are displayed in the {\bf Representation Type}. Except for our proposals, there are 13 competing methods being included and their results are adopted from a number of references~\cite{bi2020graph,sironi2018hats,lee2016training,neil2016phased}. They are H-First~\cite{orchard2015hfirst} with time spiking representation, SNN~\cite{lee2016training, neil2016phased} taking time spikings or raw events as input, HOTS and HATS~\cite{sironi2018hats} with time surfaces, several graph convolutional networks (GCNs) using graph representation (including GCN \cite{kipf2016semi}, Cheb-Conv~\cite{defferrard2016convolutional}, MoNet \cite{monti2017geometric}, GIN\cite{xu2018powerful}, G-CNN \cite{bi2019graph}, and RG-CNN \cite{bi2020graph}), as well as VGG-19~\cite{simonyan2014very}, Inception-V4 \cite{szegedy2017inception} and ResNet-50 \cite{he2016deep} with two-channel representation \cite{maqueda2018event}. 
As shown in Table \ref{tab:classification_results}, our proposed classification framework with {\bf EvRepSL} and {\bf EvRep} achieve the first and second highest classification accuracy for all datasets. The improvement is significant for the three challenging datasets, i.e., N-Caltech101, CIFAR10-DVS and ASL-DVS. {\bf Eff-Classifier} with {\bf EvRepSL} representation improves the SOTA accuracy (RG-CNN)  by 20.7\%, 23.1\% and 8.5\% for each dataset.
In addition, Eff-Classifier constraints the number of trainable parameters at 0.94 million. For comparison, VGG-19 has 144 million parameters, Inception-V4 has 43 million parameters, and ResNet-50 has 26 million parameters.

\noindent{\bf Comparison on Different Representations.}
To verify the effectiveness of the proposed representations for classification tasks, we train {\bf Eff-Classifier} with the other three compatible representations, i.e.,  two-channel~\cite{maqueda2018event}, four-channel~\cite{zhu2018ev} and voxel-grid~\cite{zhu2019unsupervised}, which are popularly used for various event-based tasks~\cite{wang2019ev, wang2021event, weng2021event, tomy2022fusing}. The classification accuracy of different representations on various datasets are reported in the last block of Table~\ref{tab:classification_results}. From the results we can observe, our proposed representations, {\bf EvRepSL} and {\bf EvRep}, achieve the first and second highest accuracy for all the datasets among the competing representations. \textcolor{HighlightColor}{We further compare the performance of \textbf{EvRepSL} with two recent event representations: the accurate and efficient event representation (which we refer to as AEER)~\cite{bai2022accurate}, and TORE~\cite{baldwin2022time}. As presented in Table~\ref{tab:recent_rep}, \textbf{EvRepSL} consistently outperforms both AEER and TORE across the majority of datasets. The only exception is the N-MNIST dataset, where EvRepSL performs marginally worse by 0.1\%. This difference is minimal and falls within the margin of error.}

% \textcolor{HighlightColor}{

\begin{table}[htbp] % htbp
\caption{\textcolor{HighlightColor}{Quantitative comparisons (Top-1 accuracy) of more recent event-stream representations.}}
\label{tab:recent_rep}
\scriptsize
% \small
\centering
\renewcommand{\arraystretch}{1.2}
\setlength{\tabcolsep}{2pt}
\textcolor{HighlightColor}{
\begin{tabular}{c|ccccc}
\hline
\hline
\textbf{Rep. Type} & \textbf{N-MNIST} &\textbf{N-Caltech101} & \textbf{N-CARS} & \textbf{CIFAR10-DVS} & \textbf{ASL-DVS}   \\
\hline
% \hline
AEER~\cite{bai2022accurate} & $0.989$ & $0.736$ & $0.923$ & $\underline{0.636}$ & $\underline{0.961}$  \\
TORE~\cite{baldwin2022time} & \boldmath${0.992}$ & $\underline{0.757}$ & $\underline{0.937}$ & ${0.607}$ & ${0.950}$ \\
EvRepSL & $\underline{0.991}$ & \boldmath$0.864$ & \boldmath$0.961$ & \boldmath$0.771$ & \boldmath$0.986$ \\
\hline
\hline
\end{tabular}
}
\end{table}

\subsection{Event-based Optical Flow Estimation}

\begin{table*}[htbp]
% \begin{tabularx*}{\textwidth}{X|l}
% \captionsetup{font=small}
\caption{{\bf Quantitative comparisons (AEE and outlier percentages) of the evaluated event-based sparse optical flow estimation methods} on \textit{indoor flying(1-3)} from MVSEC\cite{zhu2018multivehicle}. The best results are in bold while the second best results are underlined.}
\label{tab:spase_flow_results}
% \scriptsize
\small
\centering
\renewcommand{\arraystretch}{1.2}
\setlength{\tabcolsep}{2pt}
\begin{tabular}{c|c|cc|cc|cc}
\hline
\hline

% \hline
\multirow{2}{*}{\textbf{Representation Type}}    & \multirow{2}{*}{\textbf{Method Reference}}    & \multicolumn{2}{c|}{\textbf{\emph{indoor flying1}}} & \multicolumn{2}{c|}{\textbf{\emph{indoor flying2}}} & \multicolumn{2}{c}{\textbf{\emph{indoor flying3}}} \\
 & & \textbf{AEE} & \textbf{Outliers (\%)} & \textbf{AEE} & \textbf{Outliers (\%)} & \textbf{AEE} & \textbf{Outliers (\%)} \\
\hline

Two-Channel \cite{maqueda2018event}           & LIF-EV-FlowNet \cite{hagenaars2021self} & 0.71 & 1.41 & 1.44 & 12.75 & 1.16  & 9.11 \\
Four-Channel  \cite{zhu2018ev}        & EV-FlowNet \cite{zhu2018ev}     & 1.03 & 2.2  & 1.72 & 15.1  & 1.53  & 11.9 \\
Voxel-Grid    \cite{zhu2019unsupervised}         & Zhu et al. \cite{zhu2019unsupervised}     & 0.58 & \textbf{0.0}  & 1.02 & 4.0   & 0.87  & 3.0 \\
Voxel-Grid   \cite{zhu2019unsupervised}          & Paredes et al. \cite{paredes2021back} & 0.79 & 1.2  & 1.40 & 10.9  & 1.18  & 7.4 \\
EvSurface   \cite{deng2021learning}           & Deng et al. \cite{deng2021learning}    & 0.93 & 0.84 & 1.45 & 8.29  & 1.23  & 4.47  \\

Learnable MLP   \cite{gehrig2019end}        & EST \cite{gehrig2019end}            & 0.97 & 0.91 & 1.38 & 8.20  & 1.43  & 6.47 \\
Recurrent Surface   \cite{cannici2020differentiable}   & Matrix-LSTM \cite{cannici2020differentiable}    & 0.82 & 0.53 & 1.19 & 5.59  & 1.08  & 4.81 \\
Former-Latter Event Groups \cite{lee2020spike}   & Spike-FlowNet \cite{lee2020spike}  & 0.84 &   -  & 1.28 &   -   & 1.11  &  -    \\
Gaussian Weighted Polarities \cite{ding2022spatio} & STE-FlowNet \cite{ding2022spatio}    & 0.57 & 0.1  & \textbf{0.79} & \underline{1.6}   & \underline{0.72}  & \underline{1.3}    \\
\hline

Two-Channel \cite{maqueda2018event}          & \multirow{6}{*}{Eff-FlowNet}                 & 0.83 & 1.18 & 1.60 & 9.51  & 1.06  & 5.59 \\ 
Four-Channel  \cite{zhu2018ev}        &                 & 0.66 & 0.89 & 1.35 & 5.26  & 1.02  & 4.64 \\
Voxel-Grid   \cite{zhu2019unsupervised}          &                & 0.56 & 0.57 & 1.32 & 5.24  & 0.92  & 4.56 \\
EvSurface   \cite{deng2021learning}           &               & 0.58 & 0.76 & 1.44 & 6.44  & 0.99  & 4.70 \\

EvRep                  &                 & \underline{0.55} & 0.46 & 1.17 & 3.22  & 0.84  & 2.52 \\
EvRepSL                &                  & \textbf{0.42} & \underline{0.03} & \underline{0.94} & \textbf{1.48}  & \textbf{0.66}  & \textbf{0.97} \\
\hline
\hline
\end{tabular}
% \end{tabularx}
\end{table*}

\begin{table}[htbp]
% \begin{tabularx*}{\textwidth}{X|l}
% \captionsetup{font=small}
\caption{{\bf Quantitative comparisons (AEE and outlier percentages) of the evaluated event-based dense optical flow estimation methods} on \textit{indoor flying(1-3)} from MVSEC\cite{zhu2018multivehicle}. The best results are in bold while the second best results are underlined.}
\label{tab:dense_flow_results}
\scriptsize
% \small
\centering
\renewcommand{\arraystretch}{1.2}
\setlength{\tabcolsep}{2pt}
\begin{tabular}{c|c|cc|cc|cc}
\hline
\hline

% \hline
\multirow{2}{*}{\textbf{Rep. Type}}    & \multirow{2}{*}{\textbf{Method}}    & \multicolumn{2}{c|}{\textbf{\emph{ind. flying1}}} & \multicolumn{2}{c|}{\textbf{\emph{ind. flying2}}} & \multicolumn{2}{c}{\textbf{\emph{ind. flying3}}} \\
 & & \textbf{AEE} & \textbf{Out.} & \textbf{AEE} & \textbf{Out.} & \textbf{AEE} & \textbf{Out.} \\
\hline
Voxel-Grid  \cite{zhu2019unsupervised}  & Stoffregen et al. \cite{stoffregen2020reducing}  & 0.56 & 1.00 & \underline{0.66} & \textbf{1.00} & \underline{0.59} & \underline{1.00}  \\
\hline
Two-Channel \cite{maqueda2018event}   &  \multirow{6}{*}{Eff-FlowNet}                     & 0.66 & 0.82 & 0.97 & 4.06 & 0.76 & 1.85   \\
Four-Channel \cite{zhu2018ev} &                    & 0.57 & 0.61 & 0.82 & 3.05 & 0.70 & 1.79   \\
Voxel-Grid \cite{zhu2019unsupervised}   &                     & \underline{0.50} & 0.37 & 0.75 & 2.89 & 0.61 & 1.44   \\
EvSurface  \cite{deng2021learning}   &                     & 0.51 & 0.42 & 0.83 & 3.55 & 0.66 & 1.96   \\
EvRep         &                    & \underline{0.50} & \underline{0.28} & 0.77 & 2.51 & 0.61 & 1.18   \\ 
EvRepSL       &                   & \textbf{0.39} & \textbf{0.12} & \textbf{0.60} & \underline{1.80} & \textbf{0.48} & \textbf{0.81}   \\
\hline
\hline
\end{tabular}
% \end{tabularx}
\end{table}

\subsubsection{Evaluation Setup}
In this section, we evaluate and compare our proposed {\bf EvRep} and {\bf EvRepSL} with SOTA methods with various types of representation on MVSEC \cite{zhu2018multivehicle}, a dataset popularly used for optical flow estimation~\cite{zhu2019unsupervised, paredes2021back, deng2021learning, hagenaars2021self, gehrig2019end, cannici2020differentiable}. MVSEC is collected by a stereo DAVIS system paired with a LIDAR to obtain the groundtruth of optical flow~\cite{zhu2018multivehicle}. 
During the evaluation, we follow  the configuration in~\cite{gehrig2019end} for dataset split, i.e., to train the model with outdoor sequences and evaluate it with indoor sequences. We compare our methods with 10 different existing methods with 8 different types of representation. They are LIF-EV-FlowNet~\cite{hagenaars2021self} with two-channel~\cite{maqueda2018event}, EV-FlowNet~\cite{zhu2018ev} with four-channel~\cite{zhu2018ev}, Zhu et al.'s ~\cite{zhu2019unsupervised} and Paredes et al.'s approaches~\cite{paredes2021back} with voxel-grid~\cite{zhu2019unsupervised}, Deng et al.'s approach~\cite{deng2021learning} with EvSurface \cite{deng2021learning}, EST~\cite{gehrig2019end} with a learnable MLP representation~\cite{gehrig2019end}, Matrix-LSTM \cite{cannici2020differentiable} with a recurrent surface \cite{cannici2020differentiable}, Spike-FlowNet \cite{lee2020spike} with former-latter event groups \cite{lee2020spike}, STE-FlowNet \cite{ding2022spatio} with Gaussian weighted polarities \cite{ding2022spatio}, and Stoffregen et al.'s approach~\cite{stoffregen2020reducing} with voxel-grid~\cite{zhu2019unsupervised}. Except for Stoffregen et al.'s~\cite{stoffregen2020reducing}, the majority of the existing approaches focus on estimating the sparse optical flow (pixels with triggered events). In this paper, both sparse and dense estimation results of ours will be presented. To further verify the effectiveness of {\bf EvRepSL}, we also train {\bf Eff-FlowNet} with four popular representations including two-channel~\cite{hagenaars2021self}, four-channel~\cite{zhu2018ev}, voxel-grid~\cite{zhu2019unsupervised}, and EvSurface~\cite{deng2021learning}.

Average End Point Error (AEE) is used as the evaluation metric in this part and it is widely adopted for evaluating the accuracy of optical flow estimation~\cite{zhu2019unsupervised, paredes2021back, deng2021learning, hagenaars2021self, gehrig2019end, cannici2020differentiable}. AEE measures the Euclidean distance between the expected and predicted flows:

\begin{equation}
  AEE = \frac{1}{|S|}\sum_{(x,y) \in S} || F_{pred}(x, y) - F_{gt}(x, y) ||_2,
  \label{eq:aee}
\end{equation}
where $S$ represents the set of pixels annotated with flows or the set of annotated pixels with events triggered, depending on whether the estimation is dense or sparse. $F_{pred}$ and $F_{gt}$ are the predicted and expected flow vectors respectively. The subscripts $x$ and $y$ are the location pixels. Following \cite{zhu2018ev, gehrig2019end}, we additionally present the fraction of points with end point error bigger than 3 pixels and 5\% of the magnitude as outlier percentages. The smaller these two metrics indicates the higher on estimation accuracy.

\subsubsection{Evaluation Results of Optical Flow Estimation}

\noindent\textbf{Sparse Optical Flow Estimation.} As shown in Table \ref{tab:spase_flow_results}, {\bf Eff-FlowNet} with {\bf EvRepSL} achieves the best performance in 4 out of 6 indicators and the second best for the remaining two indicators compared to other methods with various representation types. In particular, for dataset \textit{indoor flying1}, the proposed method improves the accuracy of sparse optical flow estimation by a good margin compared with state-of-the-art approaches: AEE is increased by 0.15 (about 26\%) over the second best approach (STE-FlowNet~\cite{ding2022spatio}). We can also observe that {\bf EvRepSL} and {\bf EvRep} outperform the other four types of representations when using the same estimation network ({\bf Eff-FlowNet}).

\noindent\textbf{Dense Optical Flow Estimation~\footnote{A one-minute video demonstrating dense optical flow estimation with \textbf{EvRepSL} can be found in supplementary materials.}.} As demonstrated in Table \ref{tab:dense_flow_results}, {\bf Eff-FlowNet} with {\bf EvRepSL} achieves the best performance in 5 out of 6 indicators and the second best performance for the remaining one when compared with the state-of-the-art approach (Stoffregen et al \cite{stoffregen2020reducing}). In particular, for dataset \textit{indoor flying1\&3}, the proposed methods with {EvRepSL} and {EvRep} improve AEE by 0.2 (about 36\%) and 0.11 (about 19\%) respectively. Besides, the evaluation results of different types of representations with the same estimation network ({\bf Eff-FlowNet}) show that the proposed representations improve the accuracy of dense optical flow estimation effectively.

\subsection{Ablation Studies}
\noindent\textbf{Evaluating Each Channel of EvRepSL.}
\textcolor{HighlightColor}{In this section, we extend the ablation studies on the proposed representation. The comparison between EvRep and EvRepSL has already been presented in previous sections across all tasks (Tables~\ref{tab:classification_results}, ~\ref{tab:spase_flow_results}, ~\ref{tab:dense_flow_results}), where a clear performance boost is observed when incorporating the two learned channels (EvRep vs. EvRepSL). To further investigate the contribution of each channel, we progressively add new channels to the representation and report the results in Table \ref{tab:diff_channels}. The table shows that performance steadily improves as channels are added. Comparing \(\{\mathcal{E}_I, \mathcal{E}_C\}\) with EvRep highlights the effectiveness of introducing the event temporal channel \(\mathcal{E}_T\). Additionally, the table validates that both learned channels significantly contribute to the overall performance, with \(\hat{\mathcal{E}}_I^{rfd}\) having a particularly notable impact, indicating its effectiveness.}

\noindent\textbf{\textcolor{HighlightColor}{Evaluating Effectiveness of \(\mathcal{E}_I\).}}
\textcolor{HighlightColor}{To further assess the role of \(\mathcal{E}_I\), we conducted additional ablation studies on different combinations of the proposed method with and without \(\mathcal{E}_I\), as shown in Table~\ref{tab:without_ie}. The results demonstrate that the introduction of \(\mathcal{E}_I\) consistently improves performance, reinforcing its importance. Furthermore, comparing \(\{\mathcal{E}_C, \mathcal{E}_T, \mathcal{E}_I\}\) with \(EvRep + \hat{\mathcal{E}}_I^{rfd}\) shows that the learned channel \(\hat{\mathcal{E}}_I^{rfd}\) provides a marginal yet noticeable performance gain.}

% \vspace{-0.8em}

\begin{table}[htbp] % htbp
% \begin{tabularx*}{\textwidth}{X|l}
% \captionsetup{font=small}
% \captionsetup{justification=centering}
\caption{Quantitative comparisons (Top-1 accuracy) of the different selections of EvRepSL channels on five tasks.}
\label{tab:diff_channels}
\scriptsize
% \small
\centering
\renewcommand{\arraystretch}{1.2}
\setlength{\tabcolsep}{2pt}
\begin{tabular}{c|ccccc}
\hline
\hline
% \hline
\textbf{Rep.} & \textbf{N-MNIST} &\textbf{N-Caltech101} & \textbf{N-CARS} & \textbf{CIFAR10-DVS} & \textbf{ASL-DVS}   \\
% \hline
\hline

$\{\mathcal{E}_I, \mathcal{E}_C\}$ & $0.985$ & $0.732$ & $0.924$ & $0.575$ & $0.943$ \\
% \hline
EvRep & $\underline{0.990}$ & ${0.786}$ & ${0.923}$ & ${0.590}$ & ${0.977}$ \\
EvRep + $\hat{\mathcal{E}}_I^{rfd}$ & ${0.989}$ & $\underline{0.849}$ & $\underline{0.950}$ & $\underline{0.721}$ & ${0.984}$  \\
EvRep + $\hat{\theta}$ & ${0.989}$ & ${0.846}$ & ${0.941}$ & ${0.709}$ & $\underline{0.985}$ \\
EvRepSL & \boldmath$0.991$ & \boldmath$0.864$ & \boldmath$0.961$ & \boldmath$0.771$ & \boldmath$0.986$ \\

\hline
\hline
\end{tabular}
% \end{tabularx}
\end{table}

\begin{table}[htbp] % htbp
% \begin{tabularx*}{\textwidth}{X|l}
% \captionsetup{font=small}
% \captionsetup{justification=centering}
\caption{\textcolor{HighlightColor}{Quantitative comparisons (Top-1 accuracy) of the different selections of EvRepSL channels with or without $\mathcal{E}_I$.}}
\label{tab:without_ie}
\scriptsize
% \small
\centering
\renewcommand{\arraystretch}{1.2}
\setlength{\tabcolsep}{2pt}
\textcolor{HighlightColor}{
\begin{tabular}{l|ccccc}
\hline
\hline
% \hline
\textbf{Rep.} & \textbf{N-MNIST} &\textbf{N-Caltech101} & \textbf{N-CARS} & \textbf{CIFAR10-DVS} & \textbf{ASL-DVS}   \\
% \hline
\hline
$\{\mathcal{E}_C \}$ & $0.970$ & $0.707$ & $0.908$ & $0.550$ & $0.931$ \\
% \hline
$\{\mathcal{E}_C, \mathcal{E}_I \}$ & ${0.985}$ & ${0.732}$ & ${0.924}$ & ${0.575}$ & ${0.944}$ \\ \hline
$\{\mathcal{E}_C, \mathcal{E}_T \}$ & ${0.974}$ & ${0.768}$ & ${0.918}$ & ${0.578}$ & ${0.943}$  \\
$\{\mathcal{E}_C, \mathcal{E}_T, \mathcal{E}_I \}$ & ${0.990}$ & ${0.786}$ & ${0.923}$ & ${0.590}$ & ${0.977}$ \\
EvRep + $\hat{\mathcal{E}}_I^{rfd}$ & ${0.989}$ & ${0.849}$ & ${0.950}$ & ${0.721}$ & ${0.984}$  \\
\hline
\hline
\end{tabular}
}
\end{table}

\noindent\textbf{EvRepSL on Alternative Backbone.} We conduct further evaluation with different backbone network, i.e., ResNet50~\cite{he2016deep}, which is pervasively used in classification tasks. The results in Table~\ref{tab:resnet_results} demonstrate significant gain in classification accuracy when using the representation generator (EvRepSL) through five different classification tasks, which further demonstrates the generalization of our proposed EvRepSL.

% \vspace{-0.8em}

\begin{table}[htbp] % htbp
% \begin{tabularx*}{\textwidth}{X|l}
% \captionsetup{font=small}
% \captionsetup{justification=centering}
\caption{Quantitative comparisons (Top-1 accuracy) of the evaluated event-stream representations with ResNet50~\cite{he2016deep} on five tasks.}
\label{tab:resnet_results}
\scriptsize
% \small
\centering
\renewcommand{\arraystretch}{1.2}
\setlength{\tabcolsep}{2pt}
\begin{tabular}{c|ccccc}
\hline
\hline
% \hline
\textbf{Rep. Type} & \textbf{N-MNIST} &\textbf{N-Caltech101} & \textbf{N-CARS} & \textbf{CIFAR10-DVS} & \textbf{ASL-DVS}   \\
% \hline
\hline

Four-Channel & $0.953$ & $0.620$ & $0.935$ & $0.601$ & $0.969$ \\
Voxel-Grid & $0.947$ & $0.639$ & $0.934$ & $0.610$ & $0.981$  \\
EvRep & ${0.954}$ & ${0.619}$ & ${0.934}$ & ${0.605}$ & ${0.984}$ \\
EvRepSL & \boldmath$0.992$ & \boldmath$0.686$ & \boldmath$0.961$ & \boldmath$0.776$ & \boldmath$0.996$ \\

\hline
\hline
\end{tabular}
\end{table}

\subsection{\textcolor{HighlightColor}{Evaluation of Computational Efficiency}}

\textcolor{HighlightColor}{Given that low latency is a key advantage of event cameras, we optimized the processing of EvRep to minimize computational complexity. Matrix operations were employed in place of explicit loops for updating the spatial and polarity channels, allowing for faster accumulation of event data across the image grid. Additionally, event timestamps were sorted by pixel coordinates to streamline the calculation of temporal statistics. The temporal channel was further optimized using binning techniques to approximate the standard deviation of time intervals between consecutive events. By computing both the sum and squared sum of timestamp differences in matrix form, we significantly reduced the computational overhead associated with temporal dynamics. For EvRepSL, we utilized an efficient, small-scale network architecture for \textbf{RepGen}, enabling batch processing to further reduce computational cost. Following the methodology of HATS~\cite{sironi2018hats}, we report the number of events processed per second, along with the total time required to process a 100 ms sample from the N-Cars dataset. As shown in Table~\ref{tab:comp_efficiency}, EvRep can be generated at a very high speed, comparable to other methods, and even the generation of EvRepSL achieves competitive processing times.}

\begin{table}[htbp]
\caption{\textcolor{HighlightColor}{Comparison of computational efficiency in terms of processing time and speed for different methods.}}
\label{tab:comp_efficiency}
% \scriptsize
\centering
\renewcommand{\arraystretch}{1.2}
\setlength{\tabcolsep}{8pt}
\textcolor{HighlightColor}{
\begin{tabular}{l|cc}
\hline
\hline
\textbf{Method} & \textbf{Time [ms] $\downarrow$} & \textbf{Speed [kEv/s] $\uparrow$} \\
\hline
Gabor SNN~\cite{lee2016training}   & 285.95 & 14.15 \\
HOTS~\cite{lagorce2016hots}        & 157.57 & 25.68 \\
HATS~\cite{sironi2018hats}        & 7.28   & 555.74 \\
EvRep       & \boldmath$1.53$   & \boldmath$2639.65$ \\
EvRepSL     & $\underline{4.37}$   & $\underline{924.86}$ \\
\hline
\hline
\end{tabular}
}
\end{table}

\subsection{Visualization of EvRepSL}

\begin{figure}[htbp]
\vspace{-1em}
  \centering
   \includegraphics[width=0.85\linewidth]{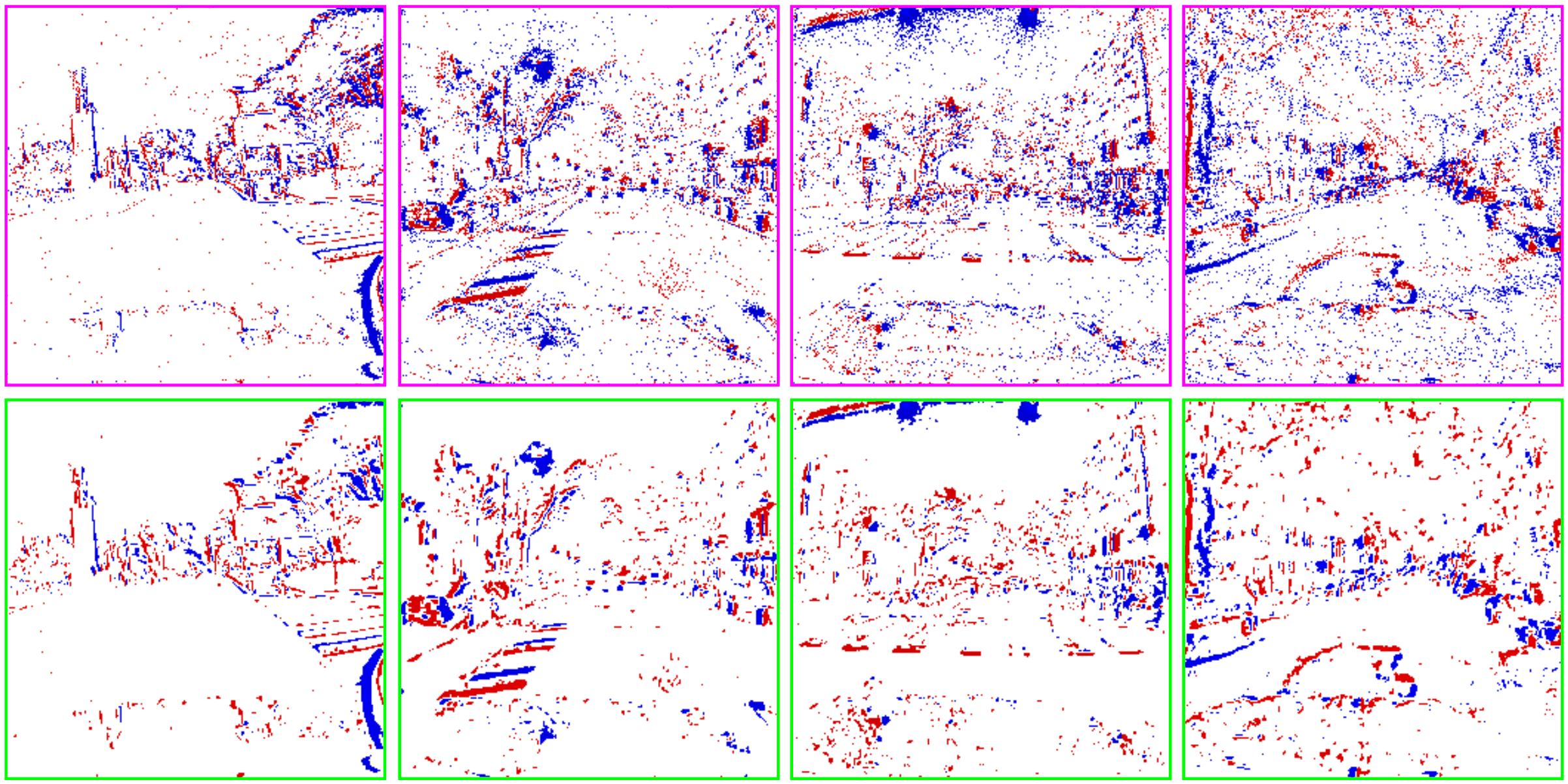}

   \caption{Visualization of $\mathcal{E}_{I}$ (first row) and $\mathcal{E}_{I}^{rfd}$ (second row)}
   \label{fig:refined_ie}
\end{figure}

Fig.~\ref{fig:refined_ie} illustrates a comparative analysis between the integral channel prior to and following the learning phase. A discernible enhancement in clarity is observed in the generated representation (depicted in the second row) when contrasted with the original (presented in the first row). The superior quality of the generated representation could potentially contribute to the exemplary performance of EvRepSL. Additional visualizations of EvRepSL channels are available in the supplementary video material.

\subsection{Discussion on Evaluation Results}

\textcolor{HighlightColor}{
The experimental results strongly demonstrate the effectiveness of our proposed method across both high-level classification and pixel-wise optical flow estimation tasks. EvRepSL outperforms its contemporaries due to two primary factors: its ability to preserve all three critical components of event data—spatial, polarity, and temporal information—and the benefits of self-supervised learning. The temporal channel \((\mathcal{E}_T)\), in particular, captures the distribution of events over time, a crucial aspect overlooked by many other representations, and this temporal information boosts performance, as seen in Table \ref{tab:diff_channels} where EvRep outperform the representation without temporal component, $\{\mathcal{E}_I, \mathcal{E}_C\}$ by a relatively large margin. Furthermore, when comparing the performance of EvRepSL to EvRep in Table \ref{tab:diff_channels}, it becomes clear that the other advantage lies in the combination of the refined integral channel and the learned contrast threshold channel, which together significantly enhance performance. This shows that the self-supervised learning stage of EvRepSL allows the RepGen model to implicitly learn valuable priors from the event streams, even without explicit labels for downstream tasks. On a finer level, EvRepSL's refined integral channel and accurate pixel-wise contrast threshold estimations contribute to its superior performance. While contrast thresholds have traditionally been viewed as constants, several studies \cite{wang2020event, brandli2014real, yang2015dynamic, kueng2016low} refute this, emphasizing the need for precise, pixel-level estimations, which further validates the advantages of our approach.
}

\textcolor{HighlightColor}{
In addition, \textbf{RepGen} functions as a representation generator, transforming raw event-streams from EvRep into the enhanced EvRepSL. After the self-supervised learning phase, the weights of \textbf{RepGen} are frozen, allowing EvRepSL to be applied directly to downstream tasks without the need for fine-tuning. This flexibility enables EvRepSL to behave like traditional representations, making it easily adaptable to a variety of tasks.} The enhancement from EvRepSL remains consistent across downstream vision tasks (according the previous experiments) and hardware configurations. Notwithstanding the diverse origins of the evaluated datasets, EvRepSL, as generated by \textbf{RepGen}, consistently enhances performance across the respective vision tasks. As delineated in Table~\ref{tab:cameras}, datasets N-MNIST, Caltech101, and CARS were sourced from the ATIS event camera~\cite{posch2010qvga}, CIFAR10-DVS from DVS128~\cite{lichtsteiner2008128}, ASL-DVS from DAVIS240C~\cite{berner2013240}, and MVSEC from DAVIS346~\cite{brandli2014real}.

Moreover, while APS frames are required for the self-learning of \textbf{RepGen}, they are not necessary for downstream applications. For example, the datasets used in the classification task mentioned above only include event-streams. Furthermore, in contrast to other event-based representation learning approaches such as EST~\cite{gehrig2019end}, the learned \textbf{RepGen} does not necessitate retraining or fine-tuning to suit different downstream tasks, thereby reducing computation and effort, making it more widely applicable.

\begin{table}[htbp]

% \captionsetup{font=small}
\caption{Event cameras used in the evaluated datasets.}
\label{tab:cameras}
\scriptsize
% \small
\centering
\renewcommand{\arraystretch}{1.2}
\setlength{\tabcolsep}{2pt}
\begin{tabular}{cccc}
\hline
\hline
\textbf{N-MNIST/Caltech101/CARS} & \textbf{CIFAR10-DVS}  & \textbf{ASL-DVS} & \textbf{MVSEC}  \\
\hline
ATIS \cite{posch2010qvga}  & DVS128 \cite{lichtsteiner2008128} & DAVIS240C \cite{berner2013240} & DAVIS346B \cite{brandli2014real} \\

\hline
\hline
\end{tabular}

\end{table}

\section{Conclusion}
\textcolor{HighlightColor}{
In this study, we introduced \textbf{EvRepSL}, a new event-stream representation developed through self-supervised learning, leveraging the relationship between APS frames and event-streams. The \textbf{RepGen} model, once trained on hybrid frame-event datasets, generates refined representations for various event-based vision tasks without requiring retraining. Our comprehensive evaluation across classification and optical flow estimation tasks on public datasets shows that \textbf{EvRepSL} consistently outperforms existing methods. Additionally, \textbf{EvRepSL} demonstrates versatility, handling different event cameras and tasks with ease, making it a strong foundation for future research in event-based vision.
}

\section*{Acknowledgments}
This work was supported in part by National Natural Science Foundation of China under Grant 62177001, Beijing Natural Science Foundation under Grant 4222003, and Shandong Provincial Natural Science Foundation Grant No. 2022HWYQ-040.

% \begin{thebibliography}{1}
{\bibliographystyle{IEEEtran}
\bibliography{main}}

% \end{thebibliography}

% \newpage

% \vfill

\end{document}